\pdfoutput=1

\documentclass[11pt]{article}

\usepackage{acl}

\usepackage{times}
\usepackage{latexsym}

\usepackage[T1]{fontenc}

\usepackage[utf8]{inputenc}
\usepackage{booktabs, multirow} 

\usepackage{microtype}

\usepackage{inconsolata}

\usepackage{graphicx}

\title{Finetuning LLMs for EvaCun 2025 Token Prediction Shared Task}

  \author{ Josef Jon, Ondřej Bojar\\
       Charles University, Faculty of Mathematics and Physics\\ 
  \texttt{surname@ufal.mff.cuni.cz}}

\begin{document}
\maketitle
\begin{abstract}
In this paper, we present our submission for the token prediction task of EvaCun 2025. Our systems are based on LLMs (Command-R, Mistral, and Aya Expanse) fine-tuned on the task data provided by the organizers. As we only possess a very superficial knowledge of the subject field and the languages of the task, we simply used the training data without any task-specific adjustments, preprocessing, or filtering. We compare 3 different approaches (based on 3 different prompts) of obtaining the predictions, and we evaluate them on a held-out part of the data.
\end{abstract}

\section{Introduction}
The EvaCun token prediction shared task focuses on missing word restoration in languages originally written in cuneiform -- Akkadian and Sumerian. The script is one of the earliest known forms of writing with a history spanning over 3000 years, evolving originally from the proto-cuneiform that was used for accounting and record keeping. One of the most challenging part of interpreting and translating Akkadian and Sumerian text is the polyvalence of cuneiform signs -- a single sign can be used as a logogram, i.e. representing a whole word (which is further complicated by the fact that one symbol can represent many different possible words, and that Akkadian texts can contain Sumerian words, even though the languages are not otherwise related), or as a syllable (one sign can represent multiple syllables) or as a determinative that denotes a semantic category of the previous word (diety, person, place, etc.). In the context of the task, our work is greatly simplified by the fact that the task data are already interpreted and transliterated into the Latin alphabet instead of being in the original cuneiform script. As we do not have any knowledge of the languages of the task in our team, we pursued a purely engineering approach of finetuning 3 different LLMs -- Aya Expanse 8B \citep{dang2024ayaexpansecombiningresearch}, Command-R v0.1 34B \citep{CommandR} and Mistral Small 3 24B \citep{Mistral} -- on the task data, with 3 slightly different formulations of the problem. We offer our solution as a baseline to be compared with the more informed and task-specific approaches.

\section{Related work}
A more focused effort in NLP for languages written in cuneiform started only recently. A basis for all future work are databases and datasets like the Electronic Text Corpus of Sumerian
Literature \cite{etcsl}, Cuneiform Digital Library Initiative, \cite{CDLI2025Home}, CuneiML \cite{Chen-2023} and
the Open Richly Annotated Cuneiform Corpus \cite{oracc}. 

\citet{simmons-etal-2024-sumtablets} created a new corpus based on these previously released datasets that pairs digital Unicode transcription of cuneiform texts with their transliterations as well as a baseline system trained on this dataset to perform this task. Similarly, \citet{akkadian_transliteration} present a method for automatic translation of Akkadian cuneiform. \citet{page-perron-etal-2017-machine} present an MT system for Sumerian with the final goal of an information retrieval pipeline for this language.  

\section{Methods}
We fine-tune autoregressive LLMs with 3 different prompts to predict the masked word. A masked language model would be a more natural choice for this task, however causal (autoregressive) language modeling is currently a more popular approach with a larger selection of pretrained models.  
\subsection{Data preprocessing and prompts}
\begin{table*}[!htp]\centering
\small
\begin{tabular}{lp{13cm}}\toprule
\textbf{Method} &\textbf{Prompt} \\\midrule
All &Fill in the missing \{language\} words, masked by the [MASK] token. Output "WORDS:" and a comma-separated list of the missing words in original \{language\}: \{masked\_document\} \\
One by one &Fill in the missing \{language\}  word masked by the [MASK] token: 
  \{masked\_document\_with\_unks\} \\
Restore &Complete the missing \{language\} words masked by the [MASK] tokens and print out the restored document: \{masked\_document\} \\
\bottomrule
\end{tabular}
\caption{Prompts for the three token prediction methods we compared.}\label{tab:prompts}

\end{table*}
The dataset provided by the organizers contains a list of tokens, each token accompanied by its document id, line number in the context of the document, the word index on that line, language, and extra information, for example, the place where the tablet was found or the type of text. We only make use of the language, word order, and document id, i.e., we do not for example split the inputs into lines or use the additional information. In each document, we mask 15\% randomly sampled words with a [MASK] token. For each document, we create at most 15 unique variants with different masks (fewer if the overall possible number of combinations for the given document length is lower).
We frame the prediction task in three different ways. The model is shown the masked document in the prompt, and it is asked to:
\begin{itemize}
    \item Produce a list containing all the original words corresponding to the masked predictions in the correct order (we call this method \textit{All} further in the text).
\item All [MASK] tokens except for one are replaced by an [UNK] token, and we ask the model to predict the original word for the single remaining [MASK] token. This is repeated for all [MASK] tokens in the masked document (\textit{One by one}).
\item We ask the model to output the full restored text of the masked document. We finetune separate models on each of these prompts (\textit{Restore}).
\end{itemize}

The specific prompts are shown in Table \ref{tab:prompts}. Our baseline approach, \textit{All}, needs the least effort for data preprocessing and training and inference compute time. However, it could suffer from error propagation due to the autoregressive nature of the inference -- the model bases the predictions on previously predicted words as well. \textit{One by one} approach could mitigate this issue, as only one masked word is predicted for each example (others remain masked). \textit{Restore} approach is based on the basic next-word prediction training objective for autoregressive models, but the decoding is complicated by the need for forcing the unmasked parts of the text and keeping the word lengths the same for whole text for both masked and unmasked versions.

\section{Experiments}
We describe the experimental setup, hyperparameters and results in this section.
\subsection{Data}
The full training data from the organizers contains 913252 tokens in 22777 documents. We set aside 1\% (227 documents) for the dev set (we filter out single word documents from the dev set). For our evaluation, we used a subset of this dev set containing 135 documents with 1500 different unique masked examples in total.
\subsection{LLM finetuning}
We finetune the pretrained models using QLoRA \cite{dettmers2023qloraefficientfinetuningquantized}. We experimented with 3 LLMs: \textit{Command-R V0.1} (4-bit quantized, \textit{CohereForAI/c4ai-command-r-v01-4bit}), \textit{Aya Expanse 8B} and \textit{Mistral Small 3  24B  Instruct} (4-bit quantized, \textit{unsloth/Mistral-Small-24B-Instruct-2501-unsloth-bnb-4bit}. We use the \texttt{transformers} \cite{wolf2020huggingfacestransformersstateoftheartnatural}, \texttt{peft} and \texttt{trl} libraries for the training. We experimented with LoRA rank sizes 8, 16, 32, 64 and 92, $\alpha=r/2$. We finetuned the models by AdamW optimizer, with warmup ratio of $0.03$ and  learning rate $lr=2e-4$. We used batch sizes 40, 36 and 35 for Aya, Command-R and Mistral models, respectively. We trained on a heterogenous cluster on a mix of Nvidia L40, A40 and H100 GPUs. We trained for a maximum of epochs, but the checkpoints we actually used for the prediction were from earlier parts of the training, as we describe in the results section.
\subsection{Results}
\setlength{\tabcolsep}{2pt}
\begin{table}[!htp]\centering
\footnotesize
\begin{tabular}{lrrrr}\toprule
\textbf{} &\textbf{Aya Expanse} &\textbf{Command-R} &\textbf{Mistral} \\\midrule
\textbf{All} &0.202 &0.209 &0.221 \\
\textbf{One by one} &0.157 &0.167 &0.205 \\
\textbf{Restore} &0.136 &0.139 &0.137 \\ \midrule
\textbf{Majority voting} &\multicolumn{3}{c}{0.269 (0.377)} \\
\textbf{Most common word} &\multicolumn{3}{c}{0.04} \\

\bottomrule
\end{tabular}
\caption{Accuracies of all combinations of models and prompts on held-out part of the task data. 
We also report top-3 accuracy for the majority voting in the parentheses. The final row shows the accuracy of predicting the most common word for the given language (based on the training data) for each masked position. } \label{tab:results}
\end{table}

\setlength{\tabcolsep}{3pt}
\begin{table}[!htp]\centering
\footnotesize
\begin{tabular}{lrrrr}\toprule
\textbf{} &\textbf{Aya Expanse}  &\textbf{Command-R} &\textbf{Mistral} \\\midrule
\textbf{All} & 6300 (0.75) & 6300 (0.67) & 5400 (0.55) \\
\textbf{One by one} & 8100 (0.15) & 5400 (0.10) &900 (0.02) \\
\textbf{Restore} &4500 (0.53) & 9000 (0.96) & 2700 (0.27) \\ 
\bottomrule
\end{tabular}
\caption{Number of updates (and the corresponding fraction of an epoch in parentheses) that the best-performing models were trained for.}\label{tab:updates}
\end{table}

\begin{figure*}
    \centering
    \includegraphics[width=0.49\linewidth]{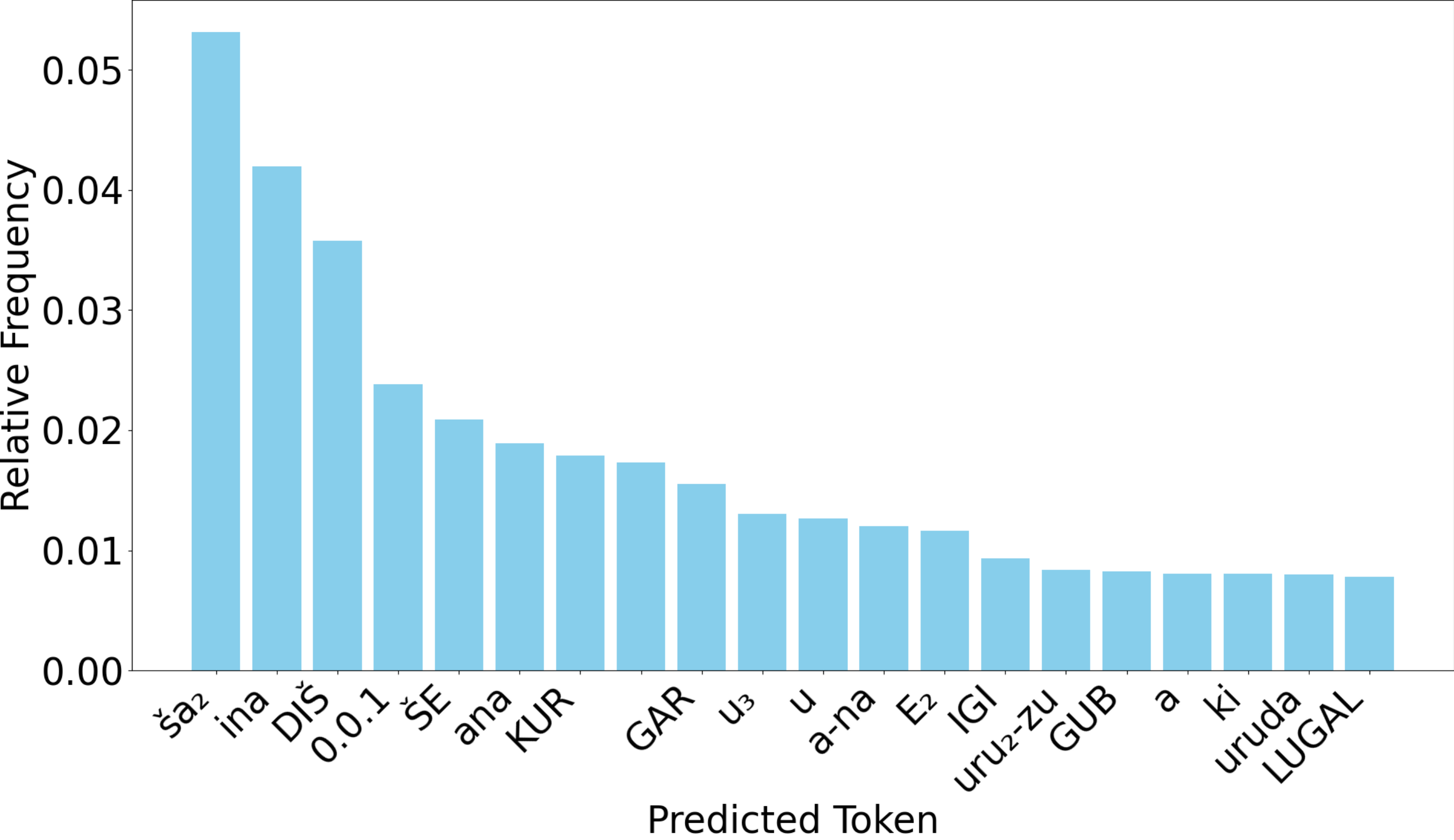}
        \includegraphics[width=0.49\linewidth]{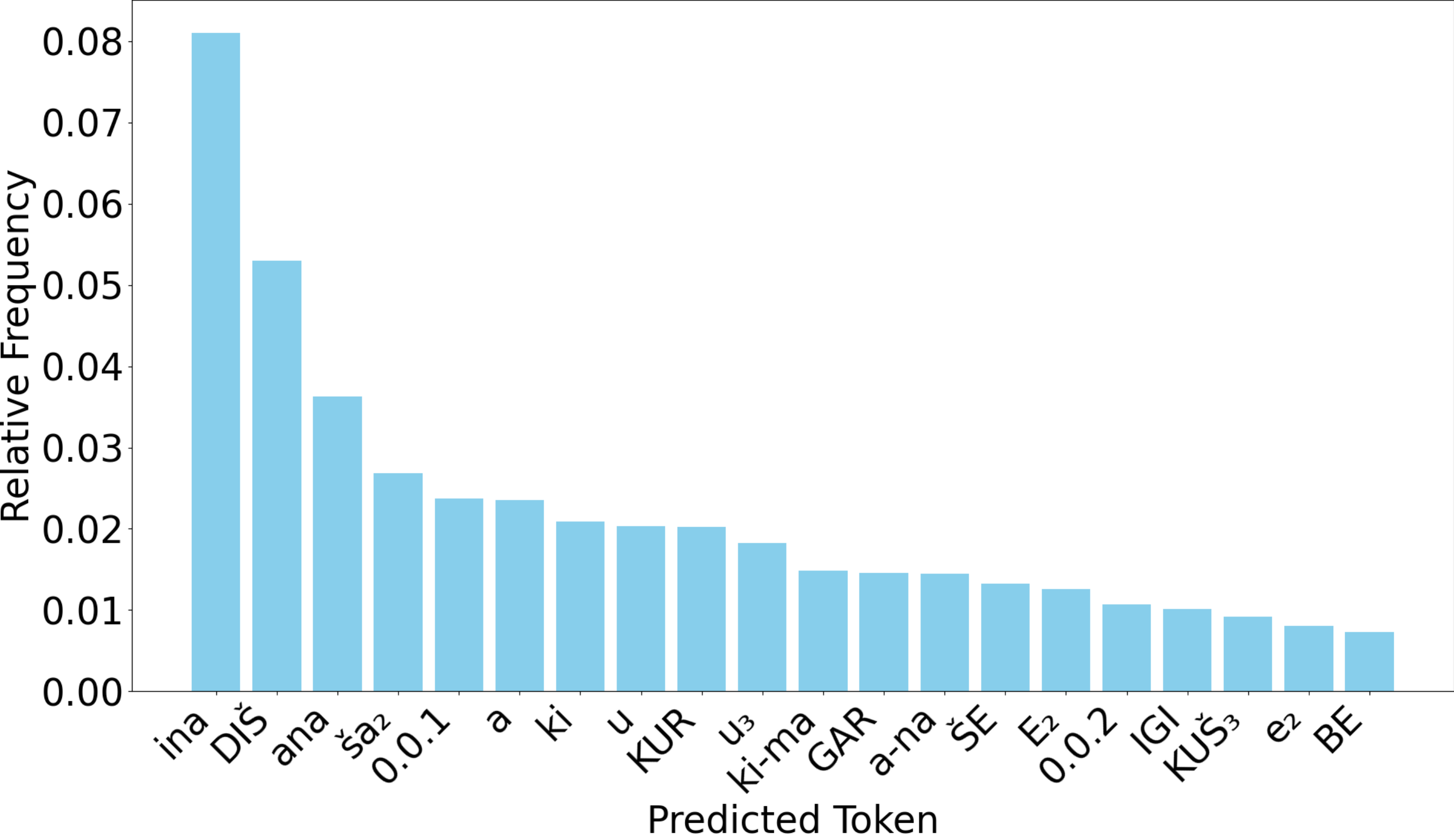} \\
    \includegraphics[width=0.49\linewidth]{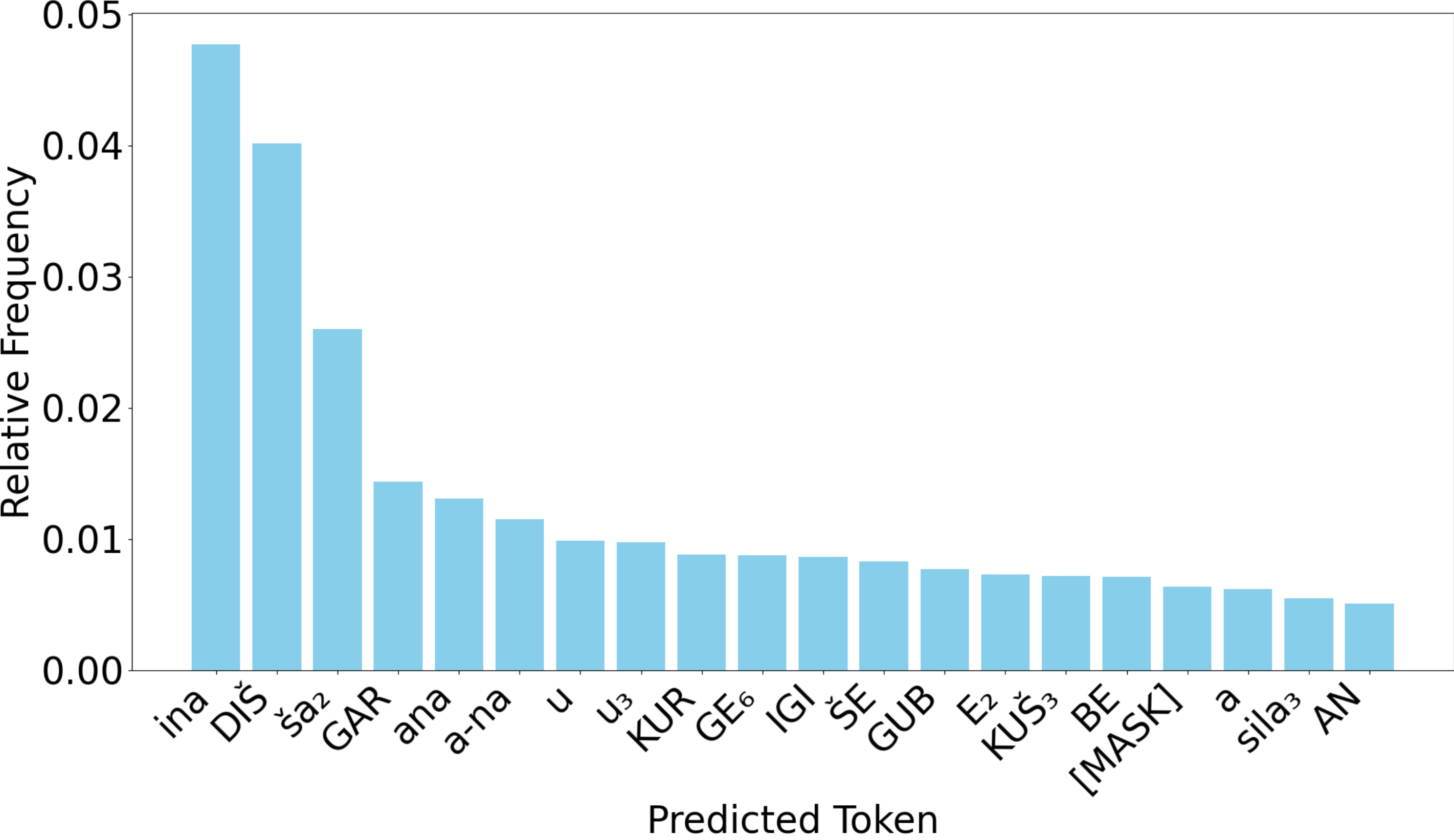}
    \includegraphics[width=0.49\linewidth]{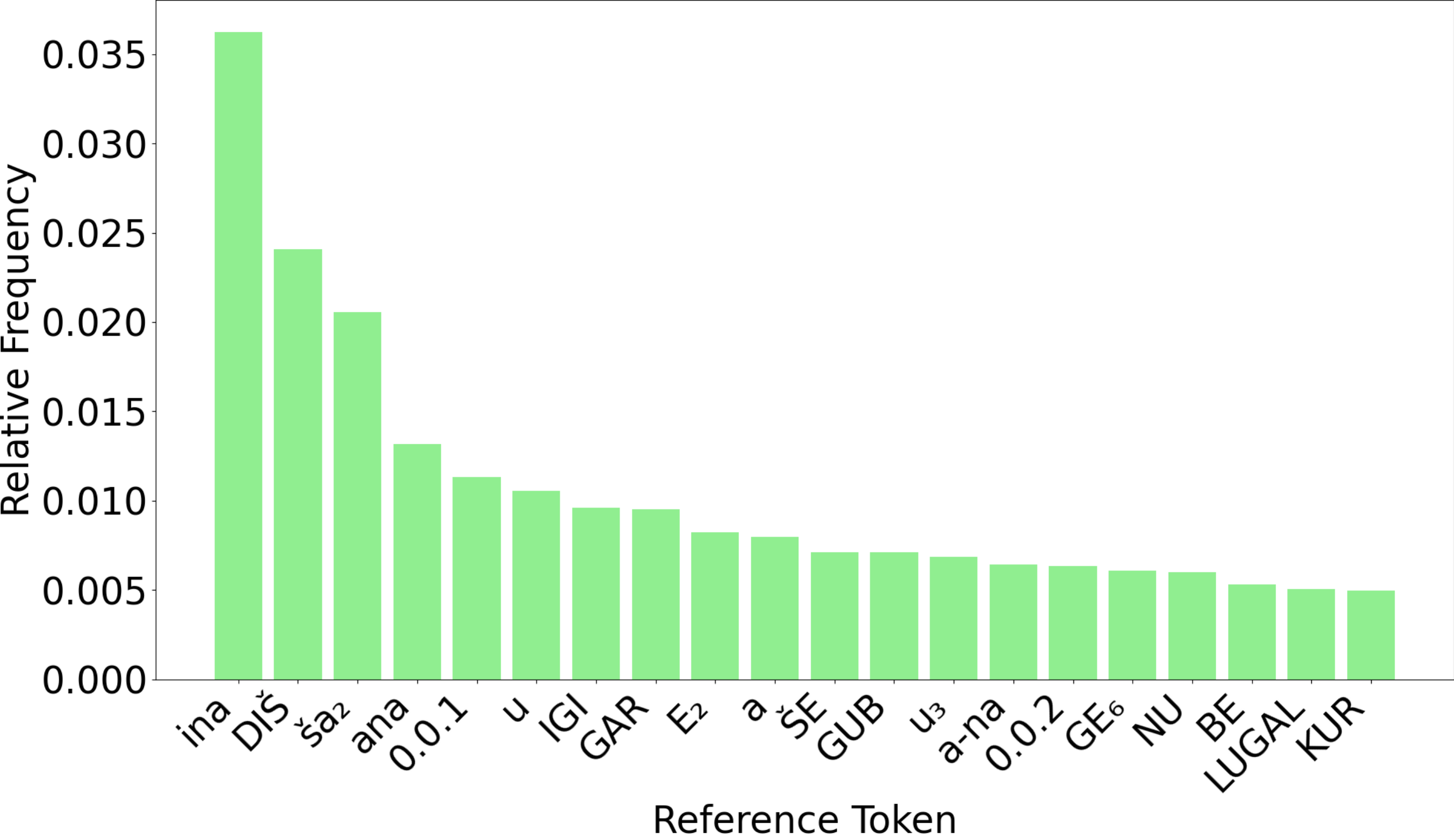}

    \caption{Relative frequencies of the top-20 generated words for masked positions by best checkpoint for each prompt (\textit{All, One by one, Restore} on top left, top right and bottom left, respectively) and of the reference masked words (bottom right).}
    \label{fig:dists}
\end{figure*} 

We sampled from the models with temperature $t=0.2$ to obtain the predictions (greedily, without the use of algorithms like beam search) and we measured the accuracy of the predictions on the held-out validation set, i.e the fraction of masked positions where the missing word was predicted correctly out of all masked positions. For the \textit{Restore} method, we generate the restored document in parts, by force decoding the known parts and only generating one complete word (possibly consisting of multiple subwords) for each [MASK] token (i.e. after force decoding the unmasked part of the document, we select the most probable subword that starts with a beginning of word symbol and generate next subwords until we reach another beginning of word subword, we discard this last subword and start with force decoding the known continuation again). We ensemble the results by majority voting, pick 60 best-performing checkpoints and select the most common prediction for each position.

We present the results of the 3 methods on the held-out validation set in Table \ref{tab:results}. Overall, finetuning the Mistral models resulted in the best accuracies. However, the differences are not large and with a different choice of hyperparameters in the finetuning, we might see different ranking. From the methods point of view, \textit{All} performed the best. Table \ref{tab:updates} shows the number of steps and a corresponding fraction of an epoch that the best-scoring checkpoints were trained on. The final line shows the majority baseline -- for each language, we only predict the most common word from the training data for all masked positions. For example, in Akkadian, the most common word is the proposition \textit{ina}, meaning \textit{in, on, onto, at, to, from} and other possible meanings in compound expressions.

We also show the relative frequencies of the top-20 predicted words by each method and of the reference words in Figure \ref{fig:dists}. We see that while the list of top 10 words is similar to the reference list for all methods, the LLMs overestimate the probability (frequency) of the most popular words. This is a common issue in text-generating models. As a result, probabilities of less common words are underestimated -- the methods generated 1567, 1217 and 3164 unique words for \textit{All, One by one} and \textit{Restore} respectively, while the reference contains 2317 unique tokens. We believe that the large number of unique tokens for the \textit{Restore} method is caused by our prediction mechanism that ensures the same word length of both the prediction and the original text but can force the selection of suboptimal predictions as a fallback.
Also, note the we did not disallow the generation of [MASK] token in the \textit{Restore} method by mistake. This negatively affects the resulting accuracy of this method. 

For the final test set submission, we ran the inference with the best 60 checkpoints on the test dataset and performed the majority voting to obtain top-3 predictions.

\section{Conclusion}
We finetuned various autoregressive LLMs on the token restoration task posed in 3 different ways. We show that the best single model can accurately predict 22.1\% of masked tokens on our held-out dev set, while by combining predictions of multiple models by voting, we can reach 26.9\% accuracy. However, there might be biases and aspects of the dataset like repetitiveness, which could lead to overestimating the real capabilities of our approach.

In the future, we plan to focus on the much more difficult task of direct translation of cuneiform script into English, either using Unicode transcriptions of the tablets, or a visual LLM to read the tablet photos directly.
\section{Acknowledgments}
This work was supported by Czech Ministry of Education, Youth and Sports (grant MŠMT OP JAK Mezisektorová spolupráce CZ.02.01.01/00/23\_020/0008518) and National Recovery Plan funded project MPO 60273/24/21300/21000 CEDMO 2.0 NPO. The computational resources were provided by Ministry of Education, Youth and Sports of the Czech Republic Project Nr. LM2023062 LINDAT/CLARIAH-CZ.

\bibliography{custom}

\appendix

\end{document}